\documentclass[10pt,twocolumn,letterpaper]{article}

\usepackage{wacv}
\usepackage{times}
\usepackage{epsfig}
\usepackage{graphicx}
\usepackage{amsmath}
\usepackage{amssymb}
\usepackage{booktabs,multirow}
\def\wacvPaperID{416} % Enter the WACV Paper ID here

\wacvalgorithmstrack   % Uncomment this line if you are submitting to the Algorithms Track.
   \wacvfinalcopy % *** Uncomment this line for the final submission

\ifwacvfinal
\usepackage[breaklinks=true,bookmarks=false]{hyperref}
\else
\usepackage[pagebackref=true,breaklinks=true,colorlinks,bookmarks=false]{hyperref}
\fi

\usepackage{bm}
\usepackage{color}
\usepackage{comment}

\definecolor{red}{rgb}{1.0, 0.0, 0.0}
\definecolor{red2}{rgb}{0.0, 0.0, 0.0}
\newcommand{\red}[1]{\textcolor{red2}{#1}}
\definecolor{TODO}{rgb}{0.0, 0.0, 1.0}
\newcommand{\TODO}[1]{\textcolor{TODO}{\\TODO: #1}}
\definecolor{IDEA}{rgb}{0.0, 0.5, 0.0}
\newcommand{\IDEA}[1]{\textcolor{IDEA}{\\IDEA: #1}}
\definecolor{Pros}{rgb}{0.0, 0.6, 0.6}
\newcommand{\Pros}[1]{\textcolor{Pros}{\\Pros: #1}}
\definecolor{Cons}{rgb}{0.8, 0.0, 0.0}
\newcommand{\Cons}[1]{\textcolor{Cons}{\\Cons: #1}}
\definecolor{CHECKPOINT}{rgb}{0.4, 0.2, 0.8}
\newcommand{\CHECKPOINT}[1]{\textcolor{CHECKPOINT}{\\CHECKPOINT: #1}}

\def\figurePath{Figures/}
\def\myfigure#1#2{\begin{figure}[htb]\centering\includegraphics*[width = \linewidth]{\figurePath#1}\caption{#2}\label{fig:#1}\end{figure}}

\def\mycfigure#1#2{\begin{figure*}[htb]\centering\includegraphics*[clip, width = \linewidth]{\figurePath#1}\caption{#2}\label{fig:#1}\end{figure*}}

\def\mysection#1#2{\section{#1}\label{sec:#2}}
\def\mysubsection#1#2{\subsection{#1}\label{sec:#2}}

\begin{document}

%%%%%%%%% TITLE
\title{Real-world Video Anomaly Detection by Extracting Salient Features in Videos}

\author{Yudai Watanabe\thanks{Current affiliation: Panasonic System Networks R\&D Lab. Co., Ltd.}\hspace{15mm}Makoto Okabe\hspace{15mm}Yasunori Harada\hspace{15mm}Naoji Kashima\\
\hspace{10mm}Shizuoka University\hspace{40mm}Chubu Electric Power Co., Inc.\\
%111-2, Itaya-cho, Naka-ku, Hamamatsu, Shizuoka, Japan\\
%20-1, Aza-kitasekiyama, Odaka-cho, Midori-ku, Nagoya, Aichi, Japan\\
{\tt\small watanabe.yudai@jp.panasonic.com, m.o@acm.org, \{Harada.Yasunori,kashima.naoji\}@chuden.co.jp}
}

\maketitle
\thispagestyle{empty}

%******************************************************
\begin{abstract}
We propose a lightweight and accurate method for detecting anomalies in videos. Existing methods used multiple-instance learning (MIL) to determine the normal/abnormal status of each segment of the video. Recent successful researches argue that it is important to learn the temporal relationships among segments to achieve high accuracy, instead of focusing on only a single segment. Therefore we analyzed the existing methods that have been successful in recent years, and found that while it is indeed important to learn all segments together, the temporal \red{orders} among them are irrelevant to achieving high accuracy. Based on this finding, we do not use the MIL framework, but instead propose a lightweight model with a self-attention mechanism to automatically extract features that are important for determining normal/abnormal from all input segments. As a result, our neural network model has 1.3\% of the number of parameters of the existing method. We evaluated the frame-level detection accuracy of our method on three benchmark datasets (UCF-Crime, ShanghaiTech, and XD-Violence) and demonstrate that our method can achieve the comparable or better accuracy than state-of-the-art methods.
\end{abstract}

%******************************************************
\mysection{Introduction}{intro}

The number of surveillance cameras in the world is increasing every year, and they are used for crime prevention in cities and for safety confirmation in factories, power plants, and other large-scale facilities. However, since it is difficult for humans to see and confirm all of these videos, there is an urgent need to develop technology that enables artificial intelligence to analyze the videos and automatically detect abnormal events on behalf of humans. Since abnormal events are rarely observed, many methods have been proposed that use only the normal state as training data, and judge whether the input video is normal or abnormal based on the criterion of how much it deviates from the learned normal state when inferring~\cite{6531615,gmm,memae,apmotion,convae}. However, these methods can only detect abnormalities based on low-level features such as differences in the appearance and velocity in the video. Therefore, recently, a method has been proposed to train an anomaly detector using a weakly supervised dataset that contains both normal and abnormal videos~\cite{real,weakly}.

In the weakly supervised dataset, each video is labeled as normal or abnormal. That is, the videos labeled as normal contain only the normal state throughout all frames. On the other hand, the videos labeled as abnormal contain a mixture of normal and abnormal frames. By using such a dataset, it is no longer necessary to label each frame in the video as normal or abnormal, thus reducing the labeling effort.

Many existing methods treated multiple consecutive frames as a single short-term segment and used multiple instance learning (MIL) to determine normal/abnormal for each segment~\cite{real}. However, recent successful methods have argued that it is important to learn the temporal relationships among segments by taking all segments from the video as input together, rather than focusing on only a single segment~\cite{weakly}. To confirm this, we analyzed these methods on a dataset of video segments randomly sorted in time. We then found that while it is indeed important to train all segments together, the temporal \red{order} among them is irrelevant for high accuracy.

Based on this finding, we propose a novel model that does not rely on MIL, but instead takes all segments as input and has a self-attention mechanism to automatically extract features important for determining normal/abnormal from them. Despite the fact that the proposed neural network has 1.3\% of the number of parameters of the existing method~\cite{weakly}, the proposed method can achieve the comparable or better accuracy than state-of-the-art methods. We report the frame-level detection accuracy on three benchmark datasets (UCF-Crime~\cite{real}, ShanghaiTech~\cite{shanghaioriginal}, and XD-Violence~\cite{xd}).

In summary, this paper has three contributions:
\begin{quote}
\begin{itemize}
\item We analyzed the existing methods and found that while it is important to learn all segments together,
the temporal \red{orders} among them are irrelevant to achieving high accuracy.
\item Based on this finding, we propose a lightweight and accurate method for video anomaly detection. Our method outperforms existing methods for the UCF-Crime dataset despite its lightweight model, a neural network with 1.3\% trainable parameters. We also show that our method outperforms existing methods for the XD-Violence dataset when the model is extended by adding a bi-directional LSTM.
\item A detailed analysis of our method was performed: visualization and observation of the attention map show that our self-attention mechanism works for extract salient features similarly to the top-$k$ strategy in existing methods. The relationship between hyperparameters and accuracy was also investigated in detail.
\end{itemize}
\end{quote}
Our method is a simple model with a self-attention mechanism and appears to be a commonplace model. However, our technical contribution is that we arrived at such a lightweight model based on our observations and analysis of existing methods and found that this model is sufficient to achieve the comparable or better accuracy than state-of-the-art methods.

%******************************************************
\mysection{Related work}{related}

In the real world, most of what we can observe are normal states, and abnormal events are rarely observed. For this reason, many anomaly detection methods have been developed using unsupervised learning approaches that learn only the normal state. In inference, the input video is judged to be normal or abnormal based on how much it deviates from the learned normal state. The normal state can be learned using \red{a set of mixture of dynamic textures models~\cite{6531615}, a space-time Markov random field (MRF) model~\cite{5206569},} Gaussian mixture models~\cite{gmm}, and sparse dictionary learning~\cite{5995434,lu2013abnormal}, etc. \red{Deep learning approaches have been also proposed, such as a method for analyzing the temporal changes in the CNN features~\cite{8354292}, autoencoder-based methods~\cite{XU2017117,memae,apmotion,convae,ionescu2019object} and GAN-based methods~\cite{8296547,anogan}, etc.} A method that uses multi-task learning has also been proposed~\cite{georgescu2021anomaly}.
However, these methods are basically only able to detect anomalies based on low-level features such as differences in the appearance and velocities of the video.

Recently, in order to develop higher-level anomaly detectors, a number of methods have been proposed to learn anomaly detectors using weakly supervised datasets containing both normal and abnormal videos~\cite{real,weakly,xd,zhang,graphconv,claws,mist,Purwanto_2021_ICCV}. Usually, to train a frame-by-frame anomaly detector, we need to label and train every frame of the video, which is expensive to label. Therefore, Sultani et al. proposed a weakly supervised dataset where not each frame but each video is labeled as normal or abnormal~\cite{real}. They also developed an anomaly detector by introducing MIL, which considers a video as a bag and selects the segment with the highest anomaly score from the bag for training.

Most of the anomaly detection methods using weakly supervised datasets are based on MIL. MIL-based methods have a problem that they have a negative impact on learning when they select normal parts of anomalous videos for learning. To cope with this problem, Zhong et al. proposed an approach that considers the weakly supervised dataset as a dataset containing incorrect labels, and uses graph convolutional neural networks to correct the incorrect labels and learn from them~\cite{graphconv}. Tian et al. proposed a top-$k$ strategy that calculates the difference in anomaly scores between segments and selects the top $k$ segments with the highest scores for training~\cite{weakly}. Sapkota et al. proposed a model that does not require the selection of the parameter $k$ by introducing distributionally robust optimization~\cite{pmlr-v130-sapkota21a}. Zaheer et al. \red{does not rely on MIL-based approach but} proposed a method for detecting anomalies using global features of the entire video and local features of each segment using the attention mechanism and per-video clustering loss~\cite{claws,https://doi.org/10.48550/arxiv.2203.13704}. These recent successful methods achieve high detection accuracy by efficiently using the features of multiple segments in the video instead of just a single segment.

\red{Recently, Zaheer et al. proposed a method for unsupervised video anomaly detection and reported significant improvements over existing unsupervised methods through experiments on the benchmark datasets (UCF-Crime and ShanghaiTech)~\cite{Zaheer_2022_CVPR}. Acsintoae et al. proposed UBnormal, a new benchmark dataset for supervised video anomaly detection consisting of virtual scenes and annotated at the pixel level~\cite{Acsintoae_2022_CVPR}.}

%******************************************************
\mysection{Method}{VC}

We propose a lightweight and accurate learning method for detecting anomalies in videos. The proposed method analyzes the entire video and automatically extracts and learns the features that are important for determining normal/abnormal.

Let $\mathcal{D}=\{(\mathbf{V}_i, y_i)\}$ be the dataset. where $\mathbf{V}_i$ is the $i$th video in the training dataset and $y_i$ is the label attached to $\mathbf{V}_i$. $y_i=\{0,1\}$, $0$ indicates normal, and $1$ indicates abnormal. In a video labeled as normal, only the normal state is recorded in all frames. On the other hand, the video labeled as abnormal contains a mixture of frames with normal and abnormal states. $\mathbf{V}_i$ is divided into $T$ segments, and each segment is converted into a $D$-dimensional feature vector $\mathbf{F}_{i,j}$ by the feature extractor: $\mathbf{F}_{i,j}$ represents the $j$th feature vector of $\mathbf{V}_i$; the feature extractor used throughout all experiments was I3D~\cite{i3d}, which has been trained on the Kineticts dataset.

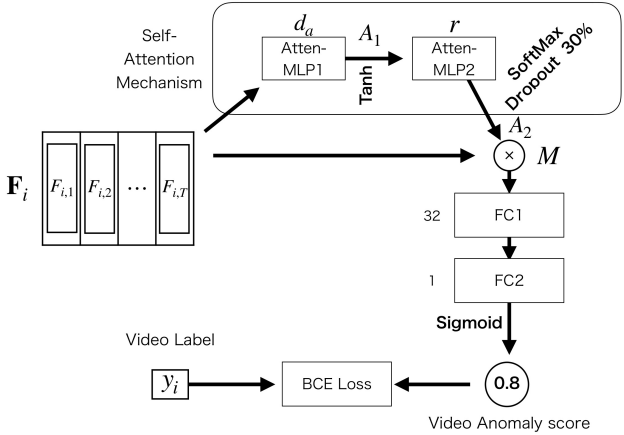
\myfigure{VideoClassifier.jpg}{An overview diagram of our method}

Our method is a simple model that consists of a self-attention mechanism and two fully connected layers (Figure~\ref{fig:VideoClassifier.jpg}). In the self-attention mechanism, the input $\mathbf{F}_i$ is transformed by Atten-MLP1 multilayer perceptron into a $d_a\times T$ matrix $\mathbf{A}_1$. At this time, the Tanh function is used for activation. $\mathbf{A}_1$ is converted to $\mathbf{A}_2$, a matrix of $r\times T$, by Atten-MLP2 multilayer perceptron. At this time, the softmax function is used for activation. In addition, a dropout regularization of 30\% is also performed.

Using the weight matrix $\mathbf{A}_2$ obtained from the self-attention mechanism, we calculate $\mathbf{M}=\mathbf{F}_i\mathbf{A}_2^t$. After reshaping $\mathbf{M}$ into a vector of $D\times r$ dimensions, it is transformed into the anomaly score through two fully connected layers FC1 (32 units) and FC2 (1 unit). The activation function of FC1 is an identity function, and that of FC2 is a sigmoid function. The binary cross entropy (BCE) function was used as the loss function.

\red{Note that our self-attention mechanism does not deal with temporal orders. As we see in the operations through Atten-MLP1 and Atten-MLP2, the vector in column $j$ of the attention map ${\bf{A}}_2$ depends only on the feature vector ${\bf{F}}_{i,j}$ in column $j$. In other words, shuffling the input feature vectors ${\bf{F}}_{i,j}$ in the temporal direction (column direction) does not change the video anomaly score, which is the output of the model.}

% ----------------------------------------
\mysubsection{Motivation of our method}{motiv}

Several existing methods take into account the temporal relationships between segments in a video~\cite{real,zhang,weakly,claws,https://doi.org/10.48550/arxiv.2203.13704}. Sultani et al. introduced a term in the loss function to impose continuity of the anomaly score, as the anomaly score should vary continuously in the video~\cite{real}.
\red{Zaheer et al. also used the similar term in their loss function~\cite{claws,https://doi.org/10.48550/arxiv.2203.13704}.}
Tian et al. introduced a multi-scale temporal network (MTN) to capture local and global temporal features~\cite{weakly}.
All of these methods achieve high accuracy in anomaly detection.
Therefore, we investigated how the mechanism for capturing the temporal relationship between segments contributes to the high accuracy.

Specifically, in $\mathbf{F_i}$, the set of feature vectors obtained from the training video $\mathbf{V_i}$, the feature vectors are ordered by default as $\{\mathbf{F}_{i,1},\mathbf{F}_{i,2},\cdots,\mathbf{F}_{i,T}\}$, but we randomly rearranged this order to create a new dataset and used it for training.
The results of the experiments using the UCF-Crime dataset~\cite{real} are shown in Table\ref{table:Shuffle}. In both methods~\cite{real,weakly}, there was no degradation in accuracy due to random reordering of feature vectors.

\begin{table}[htb]
\caption{Comparison of AUC performance using the UCF-Crime dataset~\cite{real}, where the feature vectors of each video are randomly reordered. \red{Here, we used not C3D but I3D with 10-crop augumentation for Sultani et al.'s method~\cite{real}}}
\begin{small}
\begin{center}
\label{table:Shuffle}
\begin{tabular}{ccc}
\toprule\hline
Method                                      & \hspace{8mm} Reorder \hspace{8mm} & AUC(\%) \\
\hline
\multirow{2}{*}{Sultani et al.~\cite{real}} &                                   & 81.39   \\
                                            & \checkmark                        & 81.54   \\
\hline
\multirow{2}{*}{RTFM~\cite{weakly}}         &                                   & 84.30   \\
                                            & \checkmark                        & 84.26   \\
\hline\bottomrule  
\end{tabular}
\end{center}
\end{small}
\end{table}

This result indicates that capturing the temporal \red{order} between segments does not contribute to the accuracy of the anomaly detector.

\red{The temporal smoothness term introduced by Sultani et al. minimizes the difference in anomaly scores between temporally adjacent segments~\cite{real}. However, from the above results, we infer that this term has the regularization effect of ensuring that all segments in a video have similar anomaly scores, rather than constraining the temporal order between segments. 
Zaheer et al. reported a decrease in accuracy without this term~\cite{claws,https://doi.org/10.48550/arxiv.2203.13704}, which may indicate the importance of the regularization effect that the temporal smoothness term brings. 
Although the MTN and top-$k$ strategies used in Tian et al.'s method~\cite{weakly} are more complex, and we would expect them to analyze more complex temporal relationships than just temporal order, the above results indicate that temporal order is not important.}
Based on the above observations, we hypothesized that the high anomaly detection accuracy achieved by these methods is due to the fact that they have a mechanism that allows all segments from the video to be trained together and \red{extracts salient features that are important for determining normal/abnormal.}

Our method (Figure~\ref{fig:VideoClassifier.jpg}) is designed based on the above insights.
Our method is not an MIL framework but it is a model that takes all segments in a video as input and determines whether the video is normal or abnormal.
Since we do not use MIL, the extraction of salient features can be achieved with a simple self-attention mechanism.
For the self-attention mechanism, we introduce a model inspired by Lin et al.'s method~\cite{selfattention}: 
their method targets sentence classification and can deal with variable length input.
We adopted this mechanism because we divide the video $\mathbf{V}_i$ into $T$ segments during training, but the number of segments during inference should be different dependent on the length of the input video.
Our method is accurate and also lightweight because it does not have any mechanism to capture temporal \red{orders}.

\red{Our self-attention mechanism is similar in concept to that of Claws~\cite{claws,https://doi.org/10.48550/arxiv.2203.13704}, but there are some differences between them. Claws generates the attention map twice but our method generates it once. While Claws' paper states that the purpose of introducing the self-attention mechanism is to suppress normalcy, our purpose is to extract salient features that are important in determining normality/abnormality.}

% ----------------------------------------
\mysubsection{Inference}{infer}

Let $\mathbf{V}^e$ be the video to which we want to apply our method for anomaly detection.
First, we divide $\mathbf{V}^e$ into $N$ segments $\{\mathbf{V}^e_{1},\mathbf{V}^e_{2},\cdots,\mathbf{V}^e_{N}\}$ using 16 consecutive frames as one segment.
Each segment $\mathbf{V}^e_{i}$ is converted by I3D~\cite{i3d} into a $D$-dimensional feature vector $\mathbf{F}^e_{i}$.
Let the set of feature vectors be $\mathbf{F}^e=\{\mathbf{F}^e_{1},\mathbf{F}^e_{2},\cdots,\mathbf{F}^e_{N}\}$.
Let $l$ be the split size and we divide $\mathbf{F}^e$ into $m=N/l$ bags for every $l$ segments:
\begin{eqnarray*}
\label{Inference}
\begin{split}
&\mathbf{F}^{e}=\{\mathbf{B}_1,\mathbf{B}_2,\cdots,\mathbf{B}_m\}=\\
&\{\{\mathbf{F}_{1}^e,\cdots,\mathbf{F}_{l}^e\},\{\mathbf{F}_{l+1}^e,\cdots,\mathbf{F}_{2l}^e\}\cdots\{\mathbf{F}_{N-l+1}^e,\cdots,\mathbf{F}_{N}^e\}\}.
\end{split}
\end{eqnarray*}
Next, we input $\mathbf{B}_1,\mathbf{B}_2,\cdots,\mathbf{B}_m$ one by one into our anomaly detector, and we obtain the inference result\\$\mathbf{S}^v=\{s^v_1,s^v_2,\cdots,s^v_m\}$.
$s^v_i$ is used as the anomaly score for each segment of $\{\mathbf{V}^e_{(i-1)l+1},\cdots,\mathbf{V}^e_{il}\}$.
When it is necessary to produce the results of frame-level anomaly detection, the obtained scores are assigned to all frames in each segment.

%******************************************************
\mysection{Experiments}{experiment}

To evaluate the proposed method, we conducted experiments on three weakly supervised datasets for anomaly detection: UCF-Crime dataset~\cite{real}, ShanghaiTech dataset~\cite{shanghaioriginal}, and XD-Violence dataset~\cite{xd}. We compare the detection accuracy with several existing methods.

% ----------------------------------------
\mysubsection{Dataset and evaluation measure}{}

\textbf{UCF-Crime dataset} is a large dataset of real-world surveillance video~\cite{real}. This dataset contains 13 types of anomalies. The total number of videos is 1900, and the total duration of the videos is 128 hours. 1610 of the 1900 videos are for training and 290 are for testing. Each training video is labeled as normal/abnormal on a per-video basis. In each test video, each frame is labeled as normal or abnormal.

\textbf{ShanghaiTech dataset} is a medium-sized dataset of videos captured by fixed cameras installed in the university. The dataset contains 437 videos captured by 13 fixed cameras. Of the 437 videos, 307 are normal videos and 130 are videos with anomalies. The original dataset was intended for the development of an anomaly detector based on unsupervised learning~\cite{shanghaioriginal}. However, Zhong et al. labeled each video so that it could be used as a dataset for weakly supervised learning~\cite{graphconv}. We constructed a weakly supervised dataset using the same procedure as Zhong et al.~\cite{graphconv} and conducted experiments.

\textbf{XD-Violence dataset} is a large dataset containing various types of videos, such as movies and videos from video sharing sites. The total number of videos is 4754, and the total duration of the videos is 217 hours. 2405 of the 4754 videos contain 6 types of anomalies: \textit{fighting}, \textit{shooting}, \textit{riot}, \textit{abuse}, \textit{explosion}, and \textit{car accident}. 2349 of the 4754 videos are normal. As in UCF-Crime dataset, each training video is labeled with a per-video label, and each test video is labeled with a per-frame label.

\textbf{Evaluation measure:} for UCF-Crime and ShanghaiTech datasets, we used the Area Under the Curve (AUC) with the Receiver Operating Characteristic (ROC) curve, which is calculated based on the frame-level anomaly detection accuracy, as in the existing studies~\cite{real,weakly,zhang,motionaware,graphconv,claws,xd,mist,convae,memae}; a larger AUC value indicates a more accurate anomaly detector. For XD-Violence dataset, we used Average Precision (AP) as the evaluation measure, as in the existing study~\cite{xd,weakly}; a larger AP value indicates a more accurate anomaly detector.

% ----------------------------------------
\mysubsection{Implementation detail}{}

Our method was developed and evaluated using PyTorch~\cite{pytorch}.
Radam~\cite{radam} was used as the optimization algorithm, and the learning rate was set to 0.001.
The batch size was set to 64.
As in the existing method~\cite{real}, mini-batches were created so that each mini-batch contained an equal number of normal and abnormal videos, i.e., 32 normal videos and 32 abnormal videos in our case.
The hyperparameter $T$ was set to 32.
For UCF-Crime and ShanghaiTech datasets, as in the existing method~\cite{weakly,mist}, 10-crop augmentation was performed for each video. For XD-Violence dataset, we performed 5-crop augmentation for each video as in the existing method~\cite{xd}.

% ----------------------------------------
\mysubsection{Analysis on Our Model}{videokentou}

We have experimented with adding the bi-directional long short-term memory (LSTM) to our model and removing the self-attention mechanism from our model. In the context of sentence classification, the model of Lin et al.~\cite{selfattention} achieves high classification accuracy, where the self-attention mechanism is used after the bi-directional LSTM. Inspired by this, we also extended our model by adding the bi-directional LSTM and evaluated the accuracy of anomaly detection. Specifically, our model proposed in Section~\ref{sec:VC} directly inputs $\mathbf{F}_i$ into the Atten-MLP1 (Figure~\ref{fig:VideoClassifier.jpg}). In the model with the bi-directional LSTM, $\mathbf{F}_i$ is first input to the bi-directional LSTM and the output from it is input to the Atten-MLP1. The dimension of the hidden layer of the LSTM was set to 256.

The experimental results are shown in Table~\ref{table:kentou}. For UCF-Crime and ShanghaiTech datasets, we used the features obtained by I3D, while for XD-Violence dataset, since it contains audio information, we used the features obtained by VGGish~\cite{hershey2017cnn} in addition to I3D. For UCF-Crime and ShanghaiTech datasets, the model using only the self-attention mechanism without the bi-directional LSTM (the model proposed in Section~\ref{sec:VC}) achieved the highest detection accuracy. For XD-Violence dataset, the model with both bi-directional LSTM and self-attention mechanism achieved the highest detection accuracy. This may be due to the fact that each video in XD-Violence dataset contains audio information, and the bi-directional LSTM may have worked effectively for audio information.

\begin{table}[htb]
\caption{Results of experiments on the addition of the bi-directional LSTM (BL) to our model and the removal of the self-attention mechanism (SA) from our model}
\begin{small}
\begin{center}
\label{table:kentou}
\begin{tabular}{cc|ccc}
\toprule\hline
 BL        &  SA        &  UCF-Crime     &  ShanghaiTech  &  XD-Violence   \\
\hline\hline
\checkmark &            & 81.72          & 92.90          & 75.92          \\
\hline
           & \checkmark & \textbf{84.91} & \textbf{95.72} & 75.46          \\
\hline
\checkmark & \checkmark & 83.28          & 94.06          & \textbf{82.89} \\
\hline\bottomrule
\end{tabular}
\end{center}
\end{small}
\end{table}

% ----------------------------------------
\mysubsection{Results on UCF-Crime}{ucf}

Table~\ref{table:UCF} shows the frame-level AUC performance on the UCF-Crime dataset.
The split size $l$ is set to 28.
Compared to MIST~\cite{mist} and Wu et al.~\cite{xd}, which use the same I3D RGB features, our method achieves higher detection accuracy. 
Although our method is a simple model, it achieves the highest detection accuracy 84.91\% and this is 0.61\% higher than RTFM~\cite{weakly}, which has the highest accuracy among the existing methods.

\begin{table}[htb]
\caption{Comparison of frame-level AUC performance on the UCF-Crime dataset. \textcolor{blue}{\textbf{Blue}} is the highest value and \textcolor{red}{\textbf{red}} is the second highest value}
\begin{small}
\begin{center}
\label{table:UCF}
\begin{tabular}{c|c|c|c}
\toprule\hline 
{\footnotesize Supervision}                           & Method                          & Feature Type                 & AUC(\%) \\
\hline\hline
\multirow{5}{4em}{\footnotesize One-class classifier} & SVM Baseline                    & -                            & 50.00 \\
                                                      & Conv-AE \cite{convae}           & -                            & 50.60 \\
                                                      & Lu et al. \cite{lu2013abnormal} & -                            & 65.51 \\
                                                      & BODS \cite{wang2019gods}        & -                            & 68.26 \\
                                                      & GODS \cite{wang2019gods}        & -                            & 70.46 \\
\hline
\multirow{3}{*}{\footnotesize Supervised}             & NLN \cite{nonlocal}             & NLN RGB                      & 78.9 \\
                                                      & Lin et al. \cite{background}    & C3D RGB                      & 70.1 \\
                                                      & Lin et al. \cite{background}    & NLN RGB                      & 82.0 \\
\hline
\multirow{11}{4em}{\footnotesize Weakly Supervised}   & Sultani et al. \cite{real}      & C3D RGB                      & 75.41 \\
                                                      & Zhang et al. \cite{zhang}       & C3D RGB                      & 78.66 \\
                                                      & Motion-Aware \cite{motionaware} & PWC Flow                     & 79.00 \\
                                                      & GCN-Anomaly \cite{graphconv}    & TSN RGB                      & 82.12 \\
                                                      & CLAWS Net \cite{claws}          & C3D RGB                      & 83.03 \\
                                                      & CLAWS Net+ \cite{https://doi.org/10.48550/arxiv.2203.13704} & C3D RGB   & 83.37 \\
                                                      & CLAWS Net+ \cite{https://doi.org/10.48550/arxiv.2203.13704} & 3DResNext & 84.16 \\
                                                      & Wu et al. \cite{xd}             & I3D RGB                      & 82.44 \\
                                                      & MIST \cite{mist}                & {\scriptsize I3D RGB (Fine)} & 82.30 \\
                                                      & RTFM \cite{weakly}              & C3D RGB                      & 83.28 \\
                                                      & RTFM \cite{weakly}              & I3D RGB                      & 84.30 \\
                                                      \cline{2-4} 
                                                      & Our ($d_a$=64,$r$=3)            & \multirow{2}{*}{I3D RGB}     & \textcolor{red}{\textbf{84.74}} \\
                                                      & Our ($d_a$=128,$r$=7)           &                              & \textcolor{blue}{\textbf{84.91}} \\
\hline\bottomrule
\end{tabular}
\end{center}
\end{small}
\end{table}

% ----------------------------------------
\mysubsection{Results on ShanghaiTech}{shanghai}

Table~\ref{table:shanghai} shows the frame-level AUC performance on the ShanghaiTech dataset.
The split size $l$ is set to 21.
Our detection accuracy is 95.72\% and this is 1.49\% inferior to RTFM~\cite{weakly}, which has the highest accuracy among the existing methods.
However, our method achieves the second highest value and outperforms the other existing methods.
In addition, the accuracy of more than 95\% is achieved, indicating that a sufficiently practical anomaly detector can be trained.

\begin{table}[htb]
\caption{Comparison of frame-level AUC performance on the ShanghaiTech dataset. \textcolor{blue}{\textbf{Blue}} is the highest value and \textcolor{red}{\textbf{red}} is the second highest value}
\begin{small}
\begin{center}
\label{table:shanghai}
\begin{tabular}{c|c|c|c}
\toprule\hline
{\footnotesize Supervision}                           & Method                       & Feature Type                 & AUC(\%) \\
\hline\hline
\multirow{4}{4em}{\footnotesize One-class classifier} & Conv-AE \cite{convae}        & -                            & 60.85 \\
                                                      & Frame-Pred \cite{feature}    & -                            & 73.40 \\
                                                      & Mem-AE \cite{memae}          & -                            & 71.20 \\
                                                      & VEC \cite{vec}               & -                            & 74.80 \\
\hline
\multirow{8}{4em}{\footnotesize Weakly Supervised}    & GCN-Anomaly \cite{graphconv} & TSN RGB                      & 84.44 \\
                                                      & Zhang et al. \cite{zhang}    & I3D RGB                      & 82.50 \\
                                                      & CLAWS Net \cite{claws}       & C3D RGB                      & 89.67 \\
                                                      & CLAWS Net+ \cite{https://doi.org/10.48550/arxiv.2203.13704} & C3D RGB   & 90.12 \\
                                                      & CLAWS Net+ \cite{https://doi.org/10.48550/arxiv.2203.13704} & 3DResNext & 91.46 \\
                                                      & AR-Net \cite{arnet}          & {\scriptsize I3D RGB\&Flow}  & 91.24 \\
                                                      & MIST \cite{mist}             & {\scriptsize I3D RGB (Fine)} & 94.83 \\
                                                      & RTFM \cite{weakly}           & C3D RGB                      & 91.51 \\
                                                      & RTFM \cite{weakly}           & I3D RGB                      & \textcolor{blue}{\textbf{97.21}} \\
                                                      \cline{2-4}
                                                      & Our ($d_a$=64,$r$=3)         & I3D RGB                      & \textcolor{red}{\textbf{95.72}} \\
\hline\bottomrule
\end{tabular}
\end{center}
\end{small}
\end{table}

% --------------------------------------------------------------------------------
\mysubsection{Results on XD-Violence}{}

Table~\ref{table:xd} shows the frame-level AP performance on the XD-Violence dataset.
The split size $l$ is set to 9.
When using only I3D RGB features, our method was inferior to Wu et al.~\cite{xd} and RTFM~\cite{weakly}. 
Since XD-Violence dataset contains audio information, we can also use audio features (VGGish)~\cite{hershey2017cnn} in addition to I3D RGB features as Wu et al.~\cite{xd} and Pang et al.~\cite{violance} did.
When using the audio features, as mentioned in Section~\ref{sec:videokentou}, we extend our model by adding the bi-directional LSTM in front of the self-attention mechanism, which is represented as ``Ours$\dag$'' in Table~\ref{table:xd}.
This extended model achieves the highest detection accuracy 82.89\%: this is 1.2\% higher than that of Pang et al.~\cite{violance}, which is a method dedicated to anomaly detection based on multimodal information.

\begin{table}[htb]
\caption{Comparison of frame-level AP performance on XD-Violance dataset. \textcolor{blue}{\textbf{Blue}} is the highest value and \textcolor{red}{\textbf{red}} is the second highest value. ``Ours$\dag$'' represents the model that adds bi-directional LSTM to our original model}
\begin{small}
\begin{center}
\label{table:xd}
\begin{tabular}{c|c|c|c}
\toprule\hline
{\footnotesize Supervision}                           & Method                       & Feature Type             & AP(\%) \\
\hline\hline
\multirow{2}{4em}{\footnotesize One-class classifier} & SVM baseline                 & -                        & 50.78 \\
                                                      & Hasan et al. \cite{hansen}   & -                        & 30.77 \\
\hline
\multirow{8}{4em}{\footnotesize Weakly Supervised}    & Sultani et al. \cite{real}   & C3D RGB                  & 73.20 \\
                                                      \cline{2-4} 
                                                      & Wu et al. \cite{xd}          & \multirow{3}{*}{I3D RGB} & 75.68 \\
                                                      & RTFM \cite{weakly}           &                          & 77.81 \\
                                                      & Ours ($d_a$=64,$r$=3)        &                          & 73.25 \\
                                                      \cline{2-4} 
                                                      & Wu et al. \cite{xd}          &                          & 78.64 \\
                                                      & Pang et al. \cite{violance}  & \multirow{2}{*}{I3D RGB} & \textcolor{red}{\textbf{81.69}} \\
                                                      & Ours ($d_a$=64,$r$=3)        & \multirow{2}{*}{+VGGish} & 75.46 \\
                                                      & Ours$\dag$ ($d_a$=64,$r$=3)  &                          & 79.92 \\
                                                      & Ours$\dag$ ($d_a$=128,$r$=1) &                          & \textcolor{blue}{\textbf{82.89}} \\
\hline\bottomrule
\end{tabular}
\end{center}
\end{small}
\end{table}

% ----------------------------------------
\mysubsection{Comparison of the number of \\ trainable parameters}{}

Table~\ref{table:param} shows the number of parameters that can be trained in the model for each method.
Our method is an extremely lightweight model with a much smaller number of parameters than the existing methods.
Even though the number of parameters is only 1.3\% of RTFM~\cite{weakly} when $d_a=64$ and $r=3$, our method achieves higher accuracy than RTFM~\cite{weakly} in Table~\ref{table:UCF}, and achieves a comparable high accuracy in Table~\ref{table:shanghai}.
It is also the lightest model among the existing methods even when $d_a=128$ and $r=7$, which achieves even higher accuracy.

\begin{table}[htb]
\caption{Comparison of the number of trainable parameters}
\begin{small}
\begin{center}
\label{table:param}
\begin{tabular}{cc}
\toprule\hline
Method                     & ~~~~~~Number of Parameters~~~~~~ \\
\hline
Sultani et al. \cite{real} & 2,114,113 \\
Wu et al. \cite{xd}        & 769,155 \\
RTFM \cite{weakly}         & 24,718,849 \\
\hline
Ours ($d_a=64,r=3$)        & 328,004 \\
Ours ($d_a=128,r=7$)       & 721,992 \\
\hline\bottomrule
\end{tabular}
\end{center}
\end{small}
\end{table}

% ----------------------------------------
\mysubsection{Visualization of Attention Map}{}

It was described in Section~\ref{sec:VC} that our method does not use the MIL framework and so the self-attention mechanism can be used for extracting salient features. Therefore, we visualize $\mathbf{A}_2$, the attention map, to investigate whether such salient features are really extracted. Here we visualize $\mathbf{A}_2$ for ``Buruglary030'', which is one of the training videos in the UCF-Crime dataset. Figure~\ref{fig:attention_heatmap_burglary030} shows the heatmap of $\mathbf{A}_2$ for the 50-th iteration, the 500-th iteration, and the highest detection accuracy.
Note that one iteration here means training on one mini-batch.
As the training progresses, the weights are concentrated on specific segments, indicating that the self-attention mechanism is automatically learning the segments of interest.
The fact that the weights are concentrated on multiple segments instead of one indicates that the self-attention mechanism has an effect similar to the top-$k$ strategy of RTFM~\cite{weakly}.

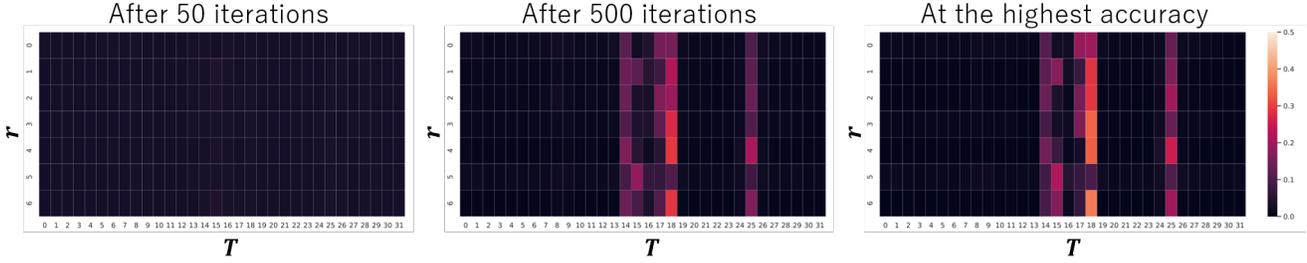
\mycfigure{attention_heatmap_burglary030}{Visualization of the attention map $\mathbf{A}_2$}

% ----------------------------------------
\mysubsection{Performance on each anomaly class}{class}

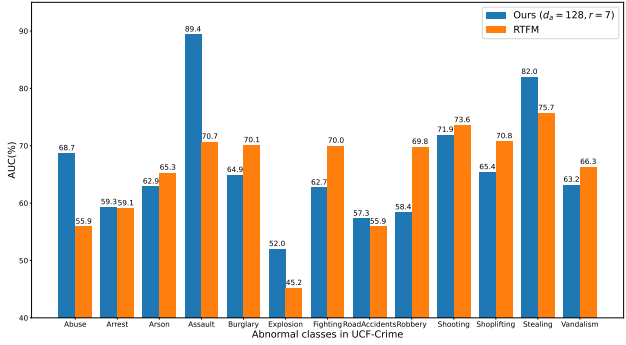
\myfigure{abnormal_class.pdf}{AUC performance by anomaly classes in UCF-Crime dataset}

Figure~\ref{fig:abnormal_class.pdf} shows the AUC performance of the proposed method for each anomaly class on the UCF-Crime dataset. Our method achieves higher or comparable detection accuracy compared to RTFM~\cite{weakly} for six classes of anomalies: Abuse, Arrest, Assault, Explosion, RoadAccidents, and Stealing. In particular, for the four classes of anomalies, Abuse, Assault, Explosion, and Stealing, the detection accuracy was improved by more than 6\%.
Since these four classes of anomalies cannot be detected without long-term analysis of the motion of objects and people, we suspect that these results indicate that the proposed method successfully captures the long-term features of the videos.
For the four classes of Arrest, Arson, RoadAccidents, and Shooting, our method performed comparably to RTFM. For the four classes of Burglary, Fighting, Robbery, and Shoplifting, our method was inferior to RTFM. Our method may be still difficult to classify instantaneous anomalies. In addition, RTFM seems to be a method that is goot at capturing features on human movement.

% ----------------------------------------
\mysubsection{Analysis on the hyperparameters $d_a$ and $r$}{}

We defined $d_a$ and $r$ as hyperparameters in Section~\ref{sec:VC}.
By changing the hyperparameters, the number of trainable parameters changes, and the detection accuracy also changes.
Therefore, using the UCF-Crime dataset, we investigated how changes in hyperparameters affect the detection accuracy.
The split size $l$ was set to 32.
The results are shown in Figure~\ref{fig:Video_classifier_parameters.pdf}.
The points surrounded in red represent the highest detection accuracy for each $d_a$.
Focusing only on $r$, if $r$ is larger than 3, detection accuracy is almost stable and its variation is small.
Focusing on the relationship between $d_a$ and $r$, if $d_a$ is increased, $r$ should also be increased to some extent to obtain better detection accuracy.
However, if $r$ is made too large, the detection accuracy will decrease.
Up to 256, better detection accuracy could be obtained by increasing $d_a$, but when $d_a$ was increased to 512, the overall detection accuracy decreased regardless of the value of $r$.
This may be due to overfitting caused by making the model too large.

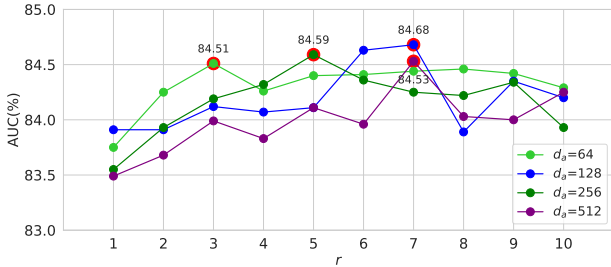
\myfigure{Video_classifier_parameters.pdf}{Relationship between hyperparameters ($d_a$ and $r$) and AUC performance on the UCF-Crime dataset}

We also investigated the influence of hyperparameters on the detection accuracy using the XD-Violence dataset.
Here we use the extended model with the bi-directional LSTM and set the split size $l$ to 32.
The results are shown in Figure~\ref{fig:xd_video_classifier_lstm_parameters.pdf}.
It can be seen that good detection accuracy is achieved when $r=1$, regardless of the value of $d_a$.
In the case of the model with the bi-directional LSTM, usage of a model that is too large for the self-attention mechanism may cause overfitting.

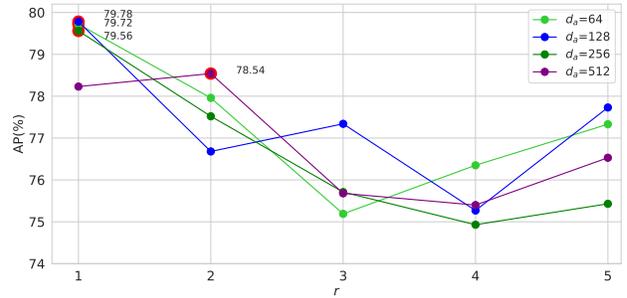
\myfigure{xd_video_classifier_lstm_parameters.pdf}{Relationship between hyperparameters ($d_a$ and $r$) and AP performance on the XD-Violence dataset}

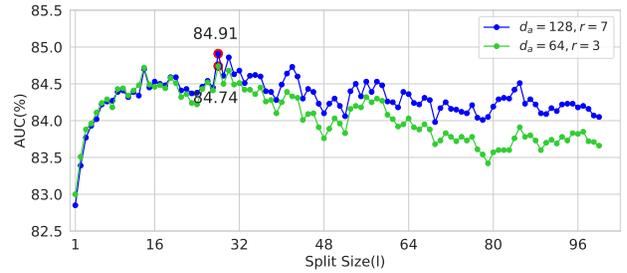
\myfigure{split_test_ucf_videoclassifier.pdf}{Relationship between the split size $l$ and AUC performance during inference on the UCF-Crime dataset}

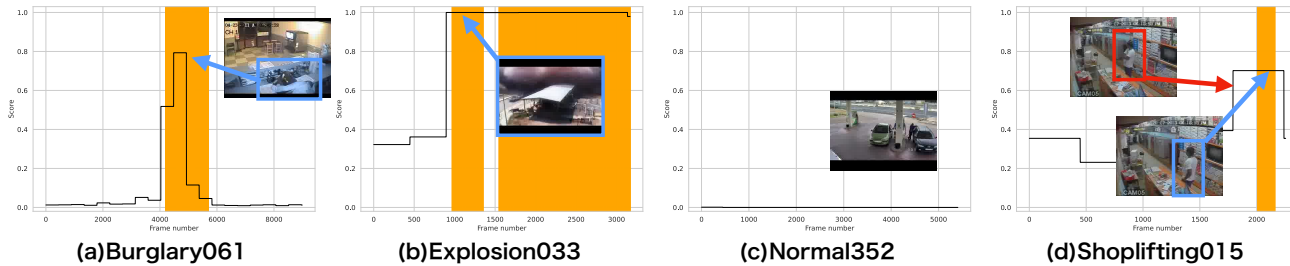
\mycfigure{visualization_video}{Visualization of the anomaly scores of our method. Black lines show the transition of anomaly scores. Orange blocks indicate ground truth. Blue arrows indicate correctly detected anomalies. Red arrows indicate incorrectly detected anomalies}

% ----------------------------------------
\mysubsection{Analysis on the split size $l$}{}

We defined the split size $l$ in Section~\ref{sec:infer}.
Since our method divides $N$ feature vectors into $m=N/l$ bags during inference, each bag contains $l$ feature vectors.
Since the detection accuracy varies depending on the split size $l$, we used the UCF-Crime dataset to investigate how the change in $l$ affects the detection accuracy.
The result for $d_a=64$ and $r=3$ and the result for $d_a=128$ and $r=7$ are shown in Figure~\ref{fig:split_test_ucf_videoclassifier.pdf}.
The points surrounded in red represent the highest detection accuracy for each hyperparameter set.
The detection accuracy is increased until $l$ is around 16.
Since our method analyzes the entire video, it requires a certain length of the video, which means that a certain split size is necessary.
On the other hand, when the split size $l$ is larger than 16, the detection accuracy is stable, indicating that our method can analyze the whole video efficiently if the video is long enough.

% ----------------------------------------
\mysubsection{Qualitative Analysis}{}

Figure~\ref{fig:visualization_video} shows the transition of anomaly scores predicted by our method for the videos of ``Burglary061'', ``Explosion033'', ``Normal352'', and ``Shoplifting015'' in the UCF-Crime datasets. The hyperparameters $d_a$ and $r$ are set to 128 and 7, respectively, and the split size $l$ is set to 16. Our method successfully detects abnormal frames when a burglar is breaking the window glass and stealing items as shown in Figure~\ref{fig:visualization_video}-a, or abnormal frames when an explosion accident happens and then dust is flying as shown in Figure~\ref{fig:visualization_video}-b. Our method correctly detects anomalies that are occurring for a long period of time. Figure~\ref{fig:visualization_video}-c shows a video at a gas station that does not contain any anomalies, and our method does not cause any false positives. Figure~\ref{fig:visualization_video}-d is an failure case. This is a video of a man shoplifting items placed on a counter. Although our method is able to detect anomalies, there are moments when false positives are detected. As described in Section~\ref{sec:class}, our method is not good at detecting short-term anomalies. For a instantaneous anomaly such as shoplifting, the frames before and after the anomaly may be identified as anomalies in addition to the moment when the anomaly actually occurred.

%******************************************************
\mysection{Conclusion}{conclusion}

We have proposed a lightweight and accurate weakly supervised learning method for anomaly detection from video.
Since MIL is not used, the extraction of salient features can be achieved with a simple self-attention mechanism.
We show that the proposed model is simple and lightweight, yet achieves the comparable or better accuracy than state-of-the-art methods.

{\small
\bibliographystyle{ieee_fullname}
\bibliography{egbib}
}

\end{document}